\documentclass[]{article}

\pdfoutput=1

\usepackage{proceed2e}

\usepackage{times}
\usepackage{amsmath}
\usepackage{amssymb}
\usepackage{graphicx}
\usepackage{color}

\usepackage{fancyhdr}
\fancypagestyle{firststyle}
{
   \fancyhf{}
   \fancyfoot[C]{\vspace{0.7em}\hrule\vspace{0.5em}\textit{Proceedings of the 32nd Conference on Uncertainty in Artificial Intelligence}, New York City, NY, USA, 2016.}
}

\DeclareMathOperator*{\argmax}{arg\,max}

\title{Bounded Rational Decision-Making in Feedforward Neural Networks}

\author{ Felix Leibfried$^{1,2,3}$ \textnormal{and} Daniel A. Braun$^{1,2}$ \\
$^1$Max Planck Institute for Intelligent Systems, T\"ubingen, Germany \\
$^2$Max Planck Institute for Biological Cybernetics, T\"ubingen, Germany\\
$^3$Graduate Training Center of Neuroscience, T\"ubingen, Germany \\
}

\begin{document}

\maketitle

\begin{abstract}
Bounded rational decision-makers transform sensory input into motor output under limited computational resources. Mathematically, such decision-makers can be modeled as information-theoretic channels with limited transmission rate. 
Here, we apply this formalism for the first time to multilayer feedforward neural networks. We derive synaptic weight update rules for two scenarios, where either each neuron is considered as a bounded rational decision-maker or the network as a whole. In the update rules, bounded rationality translates into information-theoretically motivated types of regularization in weight space.
In experiments on the MNIST benchmark classification task for handwritten digits, we show that such information-theoretic regularization successfully prevents overfitting across different architectures and attains results that are competitive with other recent techniques like dropout, dropconnect and Bayes by backprop, for both ordinary and convolutional neural networks.
\end{abstract}

\section{INTRODUCTION}

\thispagestyle{firststyle}

Intelligent systems in biology excel through their ability to flexibly adapt their behavior to changing environments so as to maximize their (expected) benefit. In order to understand such biological intelligence and to design artificial intelligent systems, a central goal is to analyze adaptive behavior from a theoretical point of view. A formal framework to achieve this goal is decision theory. An important idea, originating from the foundations of decision theory, is the principle of maximum expected utility \cite{Neumann1944}. According to the principle of maximum expected utility, an intelligent agent is formalized as a decision-maker that chooses optimal actions that maximize the expected benefit of an outcome, where the agent's benefit is quantified by a utility function.

A fundamental problem of the maximum expected utility principle is that it does not take into account computational resources that are necessary to identify optimal actions---it is for example computationally prohibitive to compute an optimal chess move because of the vast amount of potential board configurations. One way of taking computational resources into account is to study optimal decision-making under information-processing constraints \cite{Gersham2015,Simon1972}. In this study, we use an information-theoretic model of bounded rational decision-making \cite{Genewein2015,Ortega2015,Ortega2013} that has precursors in the economic literature \cite{Sims2011,Wolpert2004,Mattsson2002} and that is closely related to recent advances harnessing information theory for machine learning and perception-action systems \cite{Blundell2015,Still2014,SanchezGiraldo2013,Kappen2012,Rawlik2012,Rubin2012,Tishby2011,Friston2010,Peters2010,Still2009,Todorov2009}.

Previously, this information-theoretic bounded rationality model was applied to derive a synaptic weight update rule for a single reward-maximizing spiking neuron \cite{Leibfried2015}. It was shown that such a neuron tries to keep its firing rate close to its average firing rate, which ultimately leads to economizing of synaptic weights. Mathematically, such economizing is equivalent to a regularization that prevents synaptic weights from growing without bounds. The bounded rational weight update rule furthermore generalizes the synaptic weight update rule for an ordinary reward-maximizing spiking neuron as presented for example in \cite{Xie2004}. 
In our current work, we extend the framework of information-theoretic bounded rationality to networks of neurons, but restrict ourselves for a start to deterministic settings. In particular, we investigate two scenarios, where either each single neuron is considered as a bounded rational decision-maker or the network as a whole.

The remainder of this manuscript is organized as follows. In Section~\ref{sec:background}, we explain the information-theoretic bounded rationality model that we use. In Section~\ref{sec:neural_networks}, we apply this model to derive bounded rational synaptic weight update rules for single neurons and networks of neurons. In Section~\ref{sec:experimental_results}, we demonstrate the regularizing effect of these bounded rational weight update rules on the MNIST benchmark classification task. In Section~\ref{sec:conclusion}, we conclude.

\section{BACKGROUND ON BOUNDED RATIONAL DECISION-MAKING}
\label{sec:background}

\subsection{A FREE ENERGY PRINCIPLE FOR BOUNDED RATIONALITY}

A decision-maker is faced with the task to choose an optimal action out of a set of actions. Each action $y$ is associated with a given task-specific utility  value $U(y)$. A fully rational decision-maker picks the action $y^*$ that globally maximizes the utility function, where $y^* = \argmax_y U(y)$, assuming for notational simplicity that the global maximum is unique. Under limited computational resources however, the decision-maker may not be able to identify the globally optimal action $y^*$ which leads to the question of how limited computational resources should be quantified. In general, the decision-maker's behavior can be expressed as a probability distribution over actions $p(y)$. The basic idea of information-theoretic bounded rationality is that changes in such probability distributions are costly and necessitate computational resources. More precisely, computational resources are quantified as informational cost evoked by changing from a prior probabilistic strategy $p_0(y)$ to a posterior probabilistic strategy $p(y)$ during the deliberation process preceding the choice. Mathematically, this informational cost is given by the Kullback-Leibler divergence $D_{KL}(p(y)||p_0(y)) \leq B$ between prior and posterior strategy, where computational resources are modeled as an upper bound $B \geq 0$ \cite{Blundell2015,Ortega2015,Ortega2013,Kappen2012,Rawlik2012,Rubin2012,Friston2010,Peters2010,Still2009,Todorov2009,Wolpert2004,Mattsson2002}. Accordingly, bounded rational decision-making can be formalized by the following free energy objective
\begin{equation}
\label{eq:free_energy_objective}
\begin{split}
& p^*(y)  \\
& = \argmax_{p(y)} ~ (1-\beta) \left\langle U(y) \right\rangle_{p(y)} - \beta D_{KL}(p(y)||p_0(y))  \\ 
& = \argmax_{p(y)} ~ \left\langle  (1-\beta) U(y) - \beta \ln \frac{p(y)}{p_0(y)}  \right\rangle_{p(y)} ,
\end{split}
\end{equation}
where $\beta \in (0;1)$ controls the trade-off between expected utility and informational cost. Note that the upper bound $B$ imposed on the Kullback-Leibler divergence determines the value of $\beta$. Choosing the value of $B$ is hence equivalent to choosing the value of $\beta$.

The free energy objective in Equation~\eqref{eq:free_energy_objective} is concave with respect to $p(y)$ and the optimal solution $p^*(y)$ can be expressed in closed analytic form:
\begin{equation}
\label{eq:solution_free_energy_objective}
p^*(y) = \frac{p_0(y)\exp(\frac{1-\beta}{\beta} U(y))}{\sum_{y'} p_0(y')\exp(\frac{1-\beta}{\beta} U(y'))} .
\end{equation}
In the limit cases of none ($\beta \rightarrow 1$) and infinite  ($\beta \rightarrow 0$) resources, the optimal strategy from Equation~\eqref{eq:solution_free_energy_objective} becomes
\begin{eqnarray}
\label{eq:limit_solution_free_energy_objective_1}
\lim_{\beta \rightarrow 1} p^*(y)  = & p_0(y) ,\\
\label{eq:limit_solution_free_energy_objective_2}
\lim_{\beta \rightarrow 0} p^*(y)  = & \delta_{y y^*},
\end{eqnarray}
respectively, where $y^* = \argmax_y U(y)$ represents an action that globally maximizes the utility function. A decision-maker without any computational resources ($\beta \rightarrow 1$) sticks to its prior strategy $p_0(y)$, whereas a decision-maker that can access an arbitrarily large amount of resources ($\beta \rightarrow 0$) always picks a globally optimal action and recovers thus the fully rational decision-maker.

\subsection{A RATE DISTORTION PRINCIPLE FOR CONTEXT-DEPENDENT DECISION-MAKING}

In the face of multiple contexts, fully rational decision-making requires to find an optimal action $y$ for each environment $x$, where optimality is defined through a utility function $U(x,y)$. Bounded rational decision-making in multiple contexts means to compute multiple strategies, expressed as conditional probability distributions $p(y|x)$, under limited computational resources. Limited computational resources are modeled through an upper bound $B \geq 0$ on the expected Kullback-Leibler divergence $\left\langle D_{KL}(p(y|x)||p_0(y)) \right\rangle_{p(x)} \leq B$ between the strategies $p(y|x)$ and a common prior $p_0(y)$, averaged over all possible environments described by the distribution $p(x)$ \cite{Genewein2015,Rubin2012}. The resulting optimization problem may be formalized as
\begin{equation}
\label{eq:avg_free_energy_objective}
\begin{split}
p^*(y|x) & = \argmax_{p(y|x)} ~ (1-\beta) \left\langle U(x,y) \right\rangle_{p(x,y)}  \\
& - \beta  \left\langle D_{KL}\left(p(y|x)||p_0(y)\right)\right\rangle _{p(x)} ,
\end{split}
\end{equation}
where $\beta \in (0;1)$ governs the trade-off between expected utility and informational cost. It can be shown that the most economic prior $p_0(y)$ is given by the marginal distribution $p_0(y) = p(y) = \sum_x p(y|x)p(x)$, because the marginal distribution minimizes the expected Kullback-Leibler divergence for a given set of conditional distributions $p(y|x)$---see \cite{Tishby1999}. In this case, the expected Kullback-Leibler divergence becomes identical to the mutual information $I(x,y)$ between the environment $x$ and the action $y$ \cite{Genewein2015,Leibfried2015,Still2014,SanchezGiraldo2013,Sims2011,Tishby2011}. Accordingly, bounded rational decision-making can be formalized through the following objective
\begin{equation}
\label{eq:rate_distortion_objective}
\begin{split}
& p^*(y|x) \\
& = \argmax_{p(y|x)} ~  (1-\beta) \left\langle U(x,y) \right\rangle _{p(x,y)} - \beta I(x,y)  \\ 
& = \argmax_{p(y|x)} ~ \left\langle (1-\beta) U(x,y)  -  \beta \ln \frac{p(y|x)}{p(y)} \right\rangle_{p(x,y)} ,
\end{split}
\end{equation}
which is mathematically equivalent to the rate distortion problem from information theory \cite{Shannon1959}.

The rate distortion objective in Equation~\eqref{eq:rate_distortion_objective} is concave with respect to $p(y|x)$ but there is unfortunately no closed analytic form solution. It is however possible to express the optimal solution as a set of self-consistent equations:
\begin{eqnarray}
\label{eq:solution_rate_distortion_objective_1}
p^*(y|x) & = & \frac{p(y)\exp(\frac{1-\beta}{\beta} U(x,y) )}{\sum_{y'}p(y')\exp(\frac{1-\beta}{\beta}  U(x,y') )} , \\
\label{eq:solution_rate_distortion_objective_2}
p(y) & = & \sum_x p^*(y|x)p(x).
\end{eqnarray}
These self-consistent equations are solved by replacing $p(y)$ with an initial arbitrary distribution $q(y)$ and iterating through Equations~\eqref{eq:solution_rate_distortion_objective_1} and~\eqref{eq:solution_rate_distortion_objective_2} in an alternating fashion. This procedure is known as Blahut-Arimoto algorithm \cite{Arimoto1972,Blahut1972} and is guaranteed to converge to a global optimum \cite{Csiszar1974} presupposed that $q(y)$ does not assign zero probability mass to any $y$.

In the limit cases of none ($\beta \rightarrow 1$) and infinite ($\beta \rightarrow 0$) resources, the optimal strategy from Equations~\eqref{eq:solution_rate_distortion_objective_1} and~\eqref{eq:solution_rate_distortion_objective_2} may be expressed in closed analytic form
\begin{eqnarray}
\label{eq:limit_solution_rate_distortion_objective_1}
\lim_{\beta \rightarrow 1} p^*(y|x) = & p(y) = \delta_{y y^*} , \\
\label{eq:limit_solution_rate_distortion_objective_2}
\lim_{\beta \rightarrow 0} p^*(y|x) = & \delta_{y y^*_x},
\end{eqnarray}
where $y^* = \argmax_y \left\langle U(x,y) \right\rangle_{p(x)}$ refers to an action that globally maximizes the expected utility averaged over all possible environments, and $y_x^*$ refers to an action that globally maximizes the utility for one particular environment $x$---assuming for notational simplicity that global maxima are unique in both cases. In the absence of any computational resources ($\beta \rightarrow 1$), the decision-maker chooses the same strategy no matter which environment is encountered in order to minimize the deviation between the conditional strategies $p(y|x)$ and the average strategy $p(y)$. The decision-maker chooses however a strategy that maximizes the average expected utility. In case of access to an arbitrarily large amount of computational resources ($\beta \rightarrow 0$), the decision-maker picks the best action for each environment and recovers thus the fully rational decision-maker.

\section{THEORETICAL RESULTS: SYNAPTIC WEIGHT UPDATE RULES}
\label{sec:neural_networks}

\subsection{PARAMETERIZED STRATEGIES AND ONLINE RULES}
\label{sec:parameterized_strategies}

Computing the optimal solution to the rate distortion problem in Equation~\eqref{eq:rate_distortion_objective} with help of Equations~\eqref{eq:solution_rate_distortion_objective_1} and~\eqref{eq:solution_rate_distortion_objective_2} through the Blahut-Arimoto algorithm has two severe drawbacks. First, it requires to compute and store the conditional strategies $p(y|x)$ and the marginal strategy $p(y)$ explicitly, which is prohibitive for large environment and action spaces. And second, it requires that the decision-maker is able to evaluate the utility function for arbitrary environment-action pairs $(x,y)$, which is a plausible assumption in planning, but not in reinforcement learning where samples from the utility function can only be obtained from interactions with the environment. 

We therefore assume a parameterized form of the strategy $p_w(y|x)$, from which the decision-maker can draw samples $y$ for a given sample of the environment $x$, and optimize the rate distortion objective from Equation~\eqref{eq:rate_distortion_objective} with help of gradient ascent \cite{Leibfried2015}---also referred to as policy gradient in the reinforcement learning literature \cite{Xie2004}. Gradient ascent requires to compute the derivative of the objective function $L(w)$ with respect to the strategy parameters $w$ and to update the parameters according to the rule $w \leftarrow w + \alpha \cdot \frac{\partial}{\partial w} L(w)$ in each time step, where $\alpha > 0$ denotes the learning rate and $\frac{\partial}{\partial w} L(w)$ is defined as
\begin{equation}
\label{eq:derivative_rate_distortion_objective}
\begin{split}
& \frac{\partial}{\partial w} L(w)  = \\
&  \left\langle \left( \frac{\partial}{\partial w} \ln p_w(y|x) \right) (1-\beta) U(x,y) \right\rangle_{p_w(x,y)}  \\
 &  - \left\langle \left( \frac{\partial}{\partial w} \ln p_w(y|x) \right) \beta \ln \frac{p_w(y|x)}{p_w(y)} \right\rangle_{p_w(x,y)} .
\end{split}
\end{equation}
Note that the update rule from Equation~\eqref{eq:derivative_rate_distortion_objective} requires the computation of an
expected value over $p_w(x,y)$. This expected value can be approximated through
environment-action samples $(x,y)$ in either a batch or an online manner. For
the rest of this paper, we assume an online update rule where the agent adapts
its behavior instantaneously after each interaction with the environment in response to an immediate reward signal $U(x,y)$ as is typical for reinforcement
learning. 

Informally, the rate distortion model for bounded rational decision-making translates into a specific form of regularization that penalizes deviations of the decision-maker's instantaneous strategy $p_w(y|x)$, given the current environment $x$, from the decision-maker's mean strategy $p_w(y) = \sum_x p_w(y|x) p(x)$, averaged over all possible environments. Previously, Equation~\eqref{eq:derivative_rate_distortion_objective} was applied to a single spiking neuron that was stochastic \cite{Leibfried2015}. Here, we generalize this approach to deterministic networks of neurons that have neural input (environmental context $x$), neural output (action $y$) and a reward signal (utility $U$). We derive parameter update rules in
the style of Equation~\eqref{eq:derivative_rate_distortion_objective} that allow to adjust synaptic weights in an online fashion. In particular, we investigate two scenarios where either each single neuron
is considered as a bounded rational decision-maker or the network as a whole.

\subsection{A STOCHASTIC NEURON AS A BOUNDED RATIONAL DECISION-MAKER}

A stochastic neuron may be considered as a bounded rational decision-maker \cite{Leibfried2015}: the neuron's presynaptic input is interpreted as environmental context and the neuron's output is interpreted as action variable. The neuron's parameterized strategy corresponds to its firing behavior and is given by
\begin{equation}
\label{stochastic_neuron_strategy}
p_{\mathbf{w}}(y|\mathbf{x}) = y \cdot \rho(\mathbf{w}^\top\mathbf{x}) + (1-y) \cdot (1-\rho(\mathbf{w}^\top\mathbf{x})),
\end{equation}
where $y \in \{0,1\}$ is a binary variable reflecting the neuron's current firing state, $\mathbf{x}$ is a binary column vector representing the neuron's current presynaptic input and $\mathbf{w}$ is a real-valued column vector representing the strength of presynaptic weights. $\rho \in (0;1)$ is a monotonically increasing function denoting the neuron's current firing probability. In a similar way, the neuron's mean firing behavior can be expressed as:
\begin{equation}
\label{stochastic_neuron_mean_strategy}
p_{\mathbf{w}}(y) = y \cdot \bar{\rho}(\mathbf{w}) + (1-y) \cdot (1-\bar{\rho}(\mathbf{w})),
\end{equation}
where $\bar{\rho}(\mathbf{w}) = \sum_{\mathbf{x}} \rho(\mathbf{w}^\top\mathbf{x}) p(\mathbf{x})$ denotes the neuron's mean firing probability averaged over all possible inputs $\mathbf{x}$. The mean firing probability $\bar{\rho}(\mathbf{w})$ can be easily estimated with help of an exponential window in an online manner according to
\begin{equation}
\bar{\rho}(\mathbf{w}) \leftarrow (1-\frac{1}{\tau})\bar{\rho}(\mathbf{w}) + \frac{1}{\tau} \rho(\mathbf{w}^\top\mathbf{x}) ,
\end{equation}
where $\tau$ is a constant defining the time horizon \cite{Leibfried2015}.

Assuming a task-specific utility function $U(\mathbf{x},y)$ determining the neuron's instantaneous reward and assuming furthermore that the neuron's output $y$ does not impact the presynaptic input $\mathbf{x}$ of the next time step, the bounded rational neuron may be thought of as optimizing a rate distortion objective according to Equation~\eqref{eq:rate_distortion_objective} with gradient ascent as outlined in Section~\ref{sec:parameterized_strategies} \cite{Leibfried2015}. Equation~\eqref{eq:derivative_rate_distortion_objective} is then applicable by using the quantities
\begin{equation}
\label{eq:derivative_log_probability}
\begin{split}
& \frac{\partial}{\partial w_i} \ln p_{\mathbf{w}}(y|\mathbf{x}) = \\
 & x_i \rho'(\mathbf{w}^\top\mathbf{x}) \left( \frac{y}{\rho(\mathbf{w}^\top\mathbf{x}) } - \frac{1-y}{1-\rho(\mathbf{w}^\top\mathbf{x}) } \right),
 \end{split}
\end{equation}
and
\begin{equation}
\label{eq:mi_term}
\begin{split}
& \ln \frac{p_{\mathbf{w}}(y|\mathbf{x})}{p_{\mathbf{w}}(y)} = \\
 & y \ln \frac{\rho(\mathbf{w}^\top\mathbf{x})}{\bar{\rho}(\mathbf{w})} + (1-y) \ln \frac{1-\rho(\mathbf{w}^\top\mathbf{x}) }{1-\bar{\rho}(\mathbf{w}) }.
 \end{split}
\end{equation}
By averaging over the binary quantity $y$, a more concise weight update rule is derived \cite{Leibfried2015}:
\begin{equation}
\label{eq:derivative_rate_distortion_objective_stochastic neuron}
\begin{split}
& \frac{\partial}{\partial w_i} L(\mathbf{w})  = \\
&  \left\langle x_i \rho'(\mathbf{w}^\top\mathbf{x}) (1-\beta) \Delta U(\mathbf{x}) \right\rangle_{p(\mathbf{x})}  \\
 &  - \left\langle x_i \rho'(\mathbf{w}^\top\mathbf{x}) \beta \ln \frac{\rho(\mathbf{w}^\top\mathbf{x}) (1-\bar{\rho}(\mathbf{w}))}{\bar{\rho}(\mathbf{w}) (1-\rho(\mathbf{w}^\top\mathbf{x}))} \right\rangle_{p(\mathbf{x})} ,
\end{split}
\end{equation}
where $\Delta U(\mathbf{x}) = U(\mathbf{x},y=1) - U(\mathbf{x},y=0)$ denotes the difference in utility between firing ($y=1$) and not firing ($y=0$) for a given $\mathbf{x}$. If the conditional and marginal strategies are initialized to be roughly equal $p_{\mathbf{w}_0}(y) \approx p_{\mathbf{w}_0}(y|\mathbf{x})$, where $\mathbf{w}_0 \approx \mathbf{0}$ refers to the initial value of $\mathbf{w}$, the hyperparameter $\beta$ determines how fast the decision-maker's strategy converges. A high value of $\beta$ implies little computational resources and quick convergence due to the fact that conditional and marginal strategies are initially almost equal. On the opposite, a low value of $\beta$ indicating vast computational resources allows the decision-maker to find an optimal strategy for each environment where conditional and marginal strategies may deviate substantially.

\subsection{A DETERMINISTIC NEURON AS A BOUNDED RATIONAL DECISION-MAKER}

In a deterministic setup, the neuron's parameterized firing behavior in a small time window $\Delta t$ may be expressed through its firing rate $\phi(\mathbf{w}^\top \boldsymbol\xi)$ as:
\begin{equation}
\label{det_firing_behavior}
p_{\mathbf{w}}(y|\boldsymbol\xi)  =  y \cdot \phi(\mathbf{w}^\top\boldsymbol\xi) \Delta t + (1-y) \cdot (1-\phi(\mathbf{w}^\top\boldsymbol\xi) \Delta t),
\end{equation}
where $\boldsymbol\xi$ is a real-valued column vector indicating the presynaptic firing rates and $\phi > 0$ is a monotonically increasing function. In a similar fashion, the neuron's mean firing behavior is given by
\begin{equation}
\label{mean_firing_behavior}
p_{\mathbf{w}}(y)  = y \cdot \bar{\phi}(\mathbf{w}) \Delta t + (1-y) \cdot (1-\bar{\phi}(\mathbf{w}) \Delta t),
\end{equation}
where $\bar{\phi}(\mathbf{w}) = \sum_{\boldsymbol \xi} \phi(\mathbf{w}^\top \boldsymbol\xi) p(\boldsymbol \xi)$ refers to the neuron's mean firing rate averaged over all possible presynaptic firing rates $\boldsymbol \xi$. In accordance with the previous section, the mean firing rate $\bar{\phi}(\mathbf{w})$ can be conveniently approximated in an online manner through an exponential window with a time constant $\tau$ as:
\begin{equation}
\bar{\phi}(\mathbf{w}) \leftarrow (1-\frac{1}{\tau})\bar{\phi}(\mathbf{w}) + \frac{1}{\tau} \phi(\mathbf{w}^\top \boldsymbol \xi) .
\end{equation}
Using the quantities introduced above, we can define a mutual information rate between the presynaptic firing rates $\boldsymbol\xi$ and the instantaneous firing state of the neuron $y \in \{0;1\}$:
\begin{equation}
\label{eq:mutual_information_rate}
\begin{split}
& \lim_{\Delta t \rightarrow 0} \frac{1}{\Delta t} I(\boldsymbol\xi,y)  \\
& =  \lim_{\Delta t \rightarrow 0} \frac{1}{\Delta t} \left\langle  \sum_y p_{\mathbf{w}}(y|\boldsymbol\xi) \ln \frac{p_{\mathbf{w}}(y|\boldsymbol\xi)}{p_{\mathbf{w}}(y)} \right\rangle_{p(\boldsymbol\xi)}  \\
& =  \left\langle \phi(\mathbf{w}^\top \boldsymbol\xi) \ln \frac{\phi(\mathbf{w}^\top \boldsymbol\xi)}{\bar{\phi}(\mathbf{w})} \right\rangle_{p(\boldsymbol\xi)} .
\end{split}
\end{equation} 
A derivation of Equation~\eqref{eq:mutual_information_rate} can be found in Section~\ref{sec:mutual_info_rate_det_neu}. Assuming a rate-dependent utility function $U(\boldsymbol\xi,\phi(\mathbf{w}^\top \boldsymbol\xi))$, a deterministic neuron can be interpreted as a bounded rational decision-maker similar to Equation~\eqref{eq:rate_distortion_objective} with the following rate distortion objective
\begin{equation}
\label{eq:deterministic_rate_distortion_objective}
\begin{split}
 \mathbf{w}^* & =  \argmax_{\mathbf{w}} ~ (1-\beta) \left\langle U(\boldsymbol\xi,\phi(\mathbf{w}^\top \boldsymbol\xi)) \right\rangle _{p(\boldsymbol\xi)} \\
 &  -  \beta \lim_{\Delta t \rightarrow 0} \frac{1}{\Delta t} I(\boldsymbol\xi,y)   \\ 
 & =  \argmax_{\mathbf{w}} ~ \left\langle   (1-\beta) U(\boldsymbol\xi,\phi(\mathbf{w}^\top \boldsymbol\xi)) \right\rangle_{p(\boldsymbol\xi)} \\
 & - \left\langle \beta \phi(\mathbf{w}^\top \boldsymbol\xi) \ln \frac{\phi(\mathbf{w}^\top \boldsymbol\xi)}{\bar{\phi}(\mathbf{w})} \right\rangle_{p(\boldsymbol\xi)} .
\end{split}
\end{equation}

Optimizing the neuron's weights with gradient ascent, a similar weight update rule as in Equation~\eqref{eq:derivative_rate_distortion_objective_stochastic neuron} is derived for the deterministic case:
\begin{equation}
\label{eq:deterministic_derivative_rate_distortion_objective}
\begin{split}
& \frac{\partial}{\partial w_i} L(\mathbf{w})  =  \\
& \left\langle \xi_i \phi'(\mathbf{w}^\top \boldsymbol\xi) (1-\beta) \frac{\partial}{\partial \phi}U(\boldsymbol\xi,\phi(\mathbf{w}^\top \boldsymbol\xi)) \right\rangle_{p(\boldsymbol\xi)}\\
 & -  \left\langle \xi_i \phi'(\mathbf{w}^\top \boldsymbol\xi) \beta \ln \frac{\phi(\mathbf{w}^\top \boldsymbol\xi)}{\bar{\phi}(\mathbf{w}) } \right\rangle_{p(\boldsymbol\xi)} ,
\end{split}
\end{equation}
where $\frac{\partial}{\partial \phi}U(\boldsymbol\xi,\phi(\mathbf{w}^\top \boldsymbol\xi))$ denotes the derivative of the utility function with respect to the neuron's firing rate. The solution in Equation~\eqref{eq:deterministic_derivative_rate_distortion_objective} requires the derivative of two terms with respect to $w_i$. The derivative of the expected utility $\left\langle U(\boldsymbol\xi,\phi(\mathbf{w}^\top \boldsymbol\xi)) \right\rangle_{p(\boldsymbol\xi)}$ is straightforward, whereas the derivative of the mutual information rate $\lim_{\Delta t \rightarrow 0} \frac{1}{\Delta t} I(\boldsymbol\xi,y)$ is not so trivial and explained in more detail in Section~\ref{sec:derivative_mutual_info_rate}.

\subsection{A NEURAL NETWORK OF BOUNDED RATIONAL DETERMINISTIC NEURONS}
\label{sec:consortium}

Here, we consider a feedforward multilayer perceptron that can be imagined
to consist of individual bounded rational deterministic neurons as described in
the previous section. Assuming that all neurons aim at maximizing a global utility function while at the same time minimizing their local mutual information rate, each neuron $n$ may be interpreted as solving a deterministic rate distortion objective where the utility function
is shared among all neurons but the mutual information cost is neuron-specific:
\begin{equation}
\label{eq:global_deterministic_rate_distortion_objective}
\begin{split}
 {\mathbf{w}^n}^* & =  \argmax_{\mathbf{w}^n} ~ (1-\beta) \left\langle U(\boldsymbol\xi^{\text{in}},\mathbf{f}(\mathcal{W},\boldsymbol\xi^{\text{in}})) \right\rangle _{p(\boldsymbol\xi^{\text{in}})}  \\
 & -  \beta \lim_{\Delta t \rightarrow 0} \frac{1}{\Delta t} I(\boldsymbol\xi^n,y^n)   ,
\end{split}
\end{equation}
where ${\mathbf{w}^n}$, $\boldsymbol\xi^n$ and $y^n$ refer to the presynaptic weight vector, the presynaptic firing rates and the current firing state of neuron $n$ respectively and where $\mathcal{W}$ denotes the entirety of all weights in the whole neural network. The global utility $U(\boldsymbol\xi^{\text{in}},\mathbf{f}(\mathcal{W},\boldsymbol\xi^{\text{in}}))$ is expressed as a function of the network's input rates $\boldsymbol\xi^{\text{in}}$ and the network's output rates $\mathbf{f}(\mathcal{W},\boldsymbol\xi^{\text{in}})$.

The corresponding synaptic weight update rule for gradient ascent is similar to Equation~\eqref{eq:deterministic_derivative_rate_distortion_objective} and given by
\begin{equation}
\label{eq:global_deterministic_derivative_rate_distortion_objective}
\begin{split}
& \frac{\partial}{\partial w^n_i} L^n(\mathcal{W})  = \\
& \left\langle (1-\beta) \frac{\partial}{\partial w^n_i} U(\boldsymbol\xi^{\text{in}},\mathbf{f}(\mathcal{W},\boldsymbol\xi^{\text{in}})) \right\rangle_{p(\boldsymbol\xi^{\text{in}})} \\
& - \left\langle \beta \xi^n_i \phi'({\mathbf{w}^n}^\top \boldsymbol\xi^n) \ln \frac{\phi({\mathbf{w}^n}^\top \boldsymbol\xi^n)}{\bar{\phi}(\mathbf{w}^n) } \right\rangle_{p(\boldsymbol\xi^{\text{in}})},
\end{split}
\end{equation}
where $L^n(\mathcal{W})$ refers to the rate distortion objective of neuron $n$. The derivative of the utility function with respect to the weight $\frac{\partial}{\partial w^n_i} U(\boldsymbol\xi^{\text{in}},\mathbf{f}(\mathcal{W},\boldsymbol\xi^{\text{in}}))$ can be straightforwardly derived via ordinary backpropagation \cite{LeCun1998}.

\subsection{A DETERMINISTIC NEURAL NETWORK AS A BOUNDED RATIONAL DECISION-MAKER}
\label{sec:global_decision_maker}

While focusing on individual neurons as bounded rational decision-makers in the previous section, it is also possible to interpret an entire feedforward multilayer perceptron as one bounded rational decision-maker. To allow for this interpretation, we consider in the following the network's output rates $f_j(\mathcal{W},\boldsymbol \xi) \in (0;1)$ as the event probabilities of a categorical distribution (for example, by using a softmax activation function in the last layer). 
Importantly, the categorical distribution is considered as a bounded rational strategy 
\begin{equation}
p_\mathcal{W}(\mathbf{y}|\boldsymbol \xi) = \sum_j y_j f_j(\mathcal{W},\boldsymbol \xi) ,
\end{equation}
that generates a binary unit output vector $\mathbf{y}$ given the input rates $\boldsymbol \xi$ and the set of all weights in the entire network denoted by $\mathcal{W}$. The average bounded rational strategy is then given by
\begin{equation}
p_\mathcal{W}(\mathbf{y}) = \sum_j y_j \bar{f}_j(\mathcal{W}) ,
\end{equation}
where $\bar{f}_j(\mathcal{W})$ is the mean rate of output unit $j$ that can again be approximated in an online manner according to
\begin{equation}
\bar{f}_j(\mathcal{W}) \leftarrow (1-\frac{1}{\tau})\bar{f}_j(\mathcal{W}) + \frac{1}{\tau} f_j(\mathcal{W},\boldsymbol \xi) ,
\end{equation}
by use of an exponential window with a time constant $\tau$ in line with previous sections.

Accordingly, the informational cost can be quantified by the mutual information between $\boldsymbol \xi$ and $\mathbf{y}$:
\begin{equation}
\label{eq:mutual_information_global}
\begin{split}
I(\boldsymbol \xi,\mathbf{y}) & =  \left\langle \sum_{\mathbf{y}} p_\mathcal{W}(\mathbf{y}|\boldsymbol \xi) \ln \frac{p_\mathcal{W}(\mathbf{y}|\boldsymbol \xi)}{p_\mathcal{W}(\mathbf{y})} \right\rangle_{p(\boldsymbol \xi)} \\
& =  \left\langle \sum_{j} f_j(\mathcal{W},\boldsymbol \xi) \ln \frac{f_j(\mathcal{W},\boldsymbol \xi)}{\bar{f}_j(\mathcal{W})} \right\rangle_{p(\boldsymbol \xi)} .
\end{split}
\end{equation} 
Presupposing again a rate dependent utility function $U(\boldsymbol\xi,\mathbf{f}(\mathcal{W},\boldsymbol\xi))$, the entire deterministic network may be interpreted to solve the subsequent rate distortion objective
\begin{equation}
\label{eq:global_network_deterministic_rate_distortion_objective}
\begin{split}
 \mathcal{W}^* & =  \argmax_{\mathcal{W}} ~ (1-\beta) \left\langle U(\boldsymbol\xi,\mathbf{f}(\mathcal{W},\boldsymbol\xi)) \right\rangle _{p(\boldsymbol\xi)} \\ &  -  \beta  I(\boldsymbol\xi,\mathbf{y})  \\
  & =  \argmax_{\mathcal{W}} \ \left\langle (1-\beta) U(\boldsymbol\xi,\mathbf{f}(\mathcal{W},\boldsymbol\xi)) \right\rangle_{p(\boldsymbol\xi)} \\
  & - \left\langle \beta \sum_{j} f_j(\mathcal{W},\boldsymbol\xi) \ln \frac{f_j(\mathcal{W},\boldsymbol\xi)}{\bar{f}_j(\mathcal{W})} \right\rangle_{p(\boldsymbol\xi)} ,
\end{split}
\end{equation}

Assuming that synaptic weights are updated via gradient ascent, the following weight update rule can be derived
\begin{equation}
\label{eq:global_network_deterministic_derivative_rate_distortion_objective}
\begin{split}
& \frac{\partial}{\partial w^n_i} L(\mathcal{W})  = \\
& \left\langle (1-\beta) \sum_j \left( \frac{\partial}{\partial w^n_i} f_j(\mathcal{W},\boldsymbol\xi) \right)  \left( \frac{\partial}{\partial f_j} U(\boldsymbol\xi,\mathbf{f}(\mathcal{W},\boldsymbol\xi)) \right) \right\rangle_{p(\boldsymbol\xi)} \\
& - \left\langle \beta \sum_j \left( \frac{\partial}{\partial w^n_i} f_j(\mathcal{W},\boldsymbol\xi) \right) \ln \frac{f_j(\mathcal{W},\boldsymbol\xi)}{\bar{f}_j(\mathcal{W})}  \right\rangle_{p(\boldsymbol\xi)} ,
\end{split}
\end{equation}
where $ \frac{\partial}{\partial w^n_i}$ denotes the derivative with respect to the $i$th weight of neuron $n$, and $\frac{\partial}{\partial f_j} U(\boldsymbol\xi,\mathbf{f}(\mathcal{W},\boldsymbol\xi))$ denotes the derivative of the utility function with respect to the firing rate of the $j$th output neuron. Equation~\eqref{eq:global_network_deterministic_derivative_rate_distortion_objective} requires to differentiate two terms with respect to $w^n_i$. The derivative of the expected utility is straightforward while the derivative of the mutual information is explained in Section~\ref{sec:derivative_mutual_info_global}. 

Note that the derivative of the rate distortion objective $\frac{\partial}{\partial w^n_i} L(\mathcal{W})$ takes a convenient form which can be easily computed by extending ordinary backpropagation \cite{LeCun1998}. In ordinary backpropagation, the quantity $\frac{\partial}{\partial f_j} U(\boldsymbol\xi,\mathbf{f}(\mathcal{W},\boldsymbol\xi))$ is propagated backwards through the network. The core algorithm of ordinary backpropagation can be employed for computing $\frac{\partial}{\partial w^n_i} L(\mathcal{W})$ by simply replacing the derivative of the utility function $\frac{\partial}{\partial f_j} U(\boldsymbol\xi,\mathbf{f}(\mathcal{W},\boldsymbol\xi))$ with the more general quantity $(1-\beta)\frac{\partial}{\partial f_j} U(\boldsymbol\xi,\mathbf{f}(\mathcal{W},\boldsymbol\xi)) - \beta \ln \frac{f_j(\mathcal{W},\boldsymbol\xi)}{\bar{f}_j(\mathcal{W})}$.

\section{EXPERIMENTAL RESULTS: MNIST CLASSIFICATION}
\label{sec:experimental_results}

In our simulations, we applied both types of rate distortion regularization (the local type from Section~\ref{sec:consortium} and the global type from Section~\ref{sec:global_decision_maker}) on the MNIST benchmark classification task. In particular, we investigated in how far this information-theoretically motivated regularization subserves generalization. To this end, we trained classification on the MNIST training set, consisting of $60,000$ grayscale images of handwritten digits, and tested generalization on the MNIST test set, consisting of $10,000$ examples.
For all our simulations, we used a network with two hidden layers of rectified linear units \cite{Glorot2011} and a top layer of 10 softmax units implemented in Lua with Torch \cite{Collobert2011}. We chose as optimization criterion the negative cross entropy between the class labels and the network output \cite{Simard2003}
\begin{equation}
\label{negative_cross_entropy}
U(\boldsymbol\xi,\mathbf{f}(\mathcal{W},\boldsymbol\xi)) = \sum_j \delta_{j l(\boldsymbol\xi)} \ln f_j(\mathcal{W},\boldsymbol\xi), 
\end{equation}
where $\delta$ denotes the Kronecker delta and $\boldsymbol\xi$ the vectorized input image---note that pixels were normalized to lie in the range $[0;1]$. The variable $j \in ~ \{1, 10\}$ is an index over the network's output units and $l(\boldsymbol\xi) \in ~ \{1, 10\}$ denotes the label of image $\boldsymbol\xi$.

In order to assess the robustness of our regularizers, we performed our experiments with networks of different architectures. In particular, we used network architectures with two hidden layers and varied the number of neurons $\#_{\text{neu}} \in \{529,1024,2025,4096\}$ per hidden layer. We performed gradient ascent with a learning rate $\alpha = 0.01$ updating weights online after each training example. We trained the networks for 70 epochs where one epoch corresponded to one sweep through the entire training set. After each epoch, the learning rate decayed according to
$\alpha \leftarrow  \frac{\alpha}{1 + t \cdot \eta}$ where $t$ denotes the current epoch and $\eta = 0.002$ is a decay parameter. Weights were updated by use of a momentum $\gamma=0.9$ according to $\Delta w_i^n \leftarrow \gamma \Delta w_i^n + (1-\gamma) \frac{\partial}{\partial w_i^n} L(\mathcal{W})$ and were randomly initialized in the range $(-(\#_\text{in}(n))^{-0.5}; (\#_\text{in}(n))^{-0.5})$ with help of a uniform distribution at the beginning of the simulation where $\#_\text{in}(n)$ denotes the number of inputs to neuron $n$. Each non-input neuron had an additional bias weight that was initialized in the same way as the presynaptic weights of that neuron. Rate distortion regularization required furthermore to compute the mean firing rate $\bar{\phi}(\mathbf{w}^n)$ of individual neurons $n$ through an exponential window in an online fashion with a time constant $\tau = 1000$. In order to ensure numerical stability when using rate distortion regularization, terms of the form $\ln \frac{\phi({\mathbf{w}^n}^\top \boldsymbol\xi^n)}{\bar{\phi}(\mathbf{w}^n)}$ in the weight update rules were computed according to $\ln \max\{\phi({\mathbf{w}^n}^\top \boldsymbol\xi^n), \varepsilon\} - \ln \max\{\bar{\phi}(\mathbf{w}^n),\varepsilon\}$ with $\varepsilon = 2.22 \cdot 10^{-16}$.

\begin{figure*}[ht]
\vspace{1in}
\begin{center}
\includegraphics[width=1\linewidth]{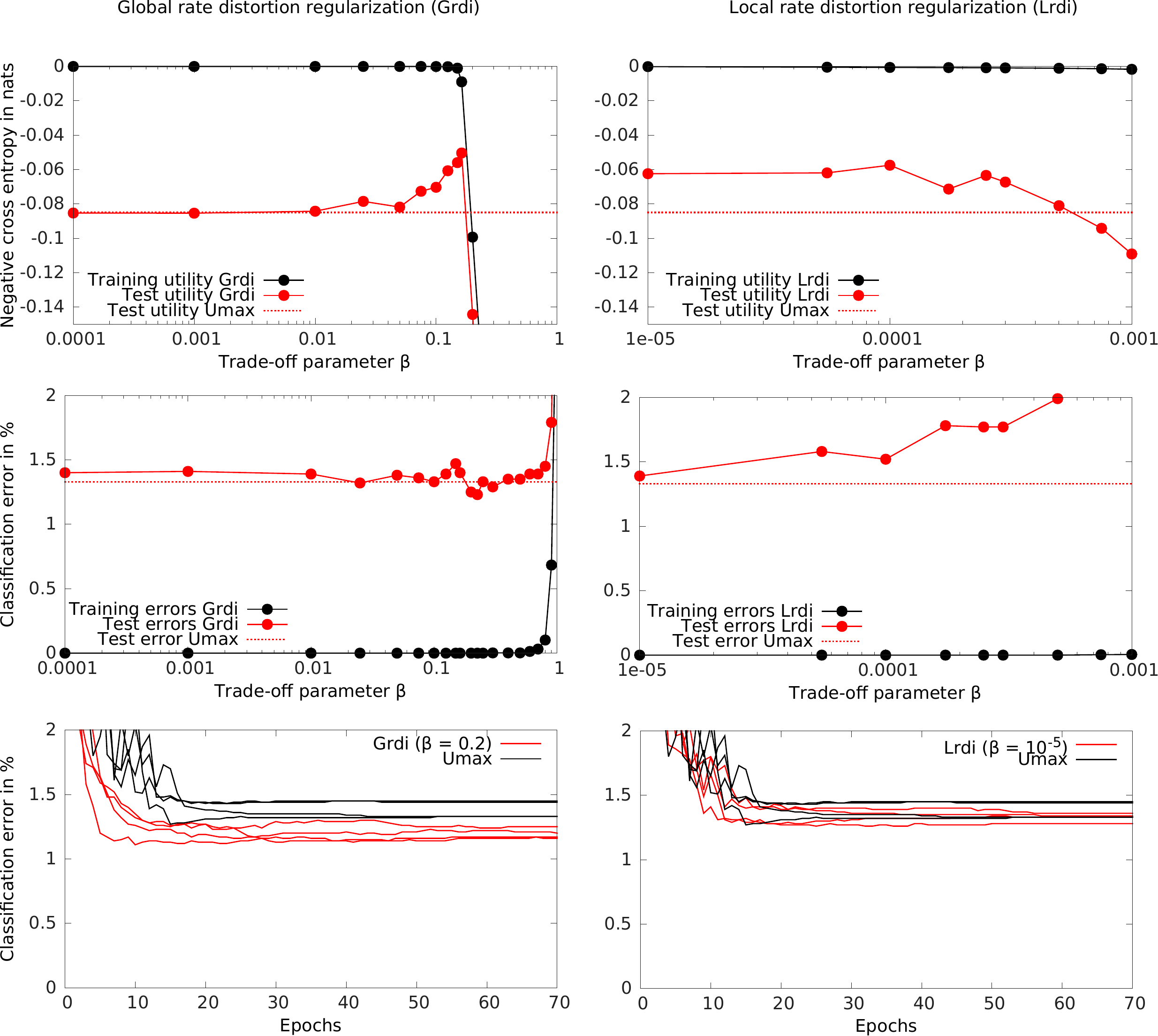}
\caption{Performance on MNIST in the Permutation Invariant Setup. The left column refers to analyses with global rate distortion regularization (Grdi) and the right column to analyses with local rate distortion regularization (Lrdi). 
The \emph{upper and middle panels} show the results of the pilot studies  on the smallest network architecture. 
Trajectories in the upper panels illustrate the expected utility (the negative cross entropy) after 50 epochs of training for different values of $\beta$---black solid lines reflect the expected utility on the training set, red solid lines reflect the expected utility on the test set and red dashed horizontal lines reflect the expected utility on the test set in ordinary utility maximization (Umax, $\beta=0$). The middle panels show classification errors instead of utility values. In the Grdi case, the negative cross entropy drops sharply for larger betas, because the regularization drives the output rates towards a flatter distribution, even though the mode of the distribution is maintained, which allows for robust performance in terms of classification error.
The \emph{lower panels} show the results of our final simulations with four different network architectures and fixed $\beta$-values. The plots compare the development of the test set error over epochs between ordinary utility maximization (black trajectories, Umax) and rate distortion regularization (red trajectories, Grdi with $\beta=0.2$ and Lrdi with $\beta=10^{-5}$ respectively). Each trajectory corresponds to one of the four different network architectures.}
\label{fig:results}
\end{center}
\end{figure*}
 
 To find optimal values for the rate distortion trade-off parameter $\beta$, we conducted pilot studies with small networks comprising 529 neurons per hidden layer that were trained for only 50 epochs on the MNIST training set according to the aforementioned training scheme and subsequently evaluated on the MNIST test set. While this might induce overfitting of $\beta$ on the test set in the small networks, we used the same $\beta$-values as a heuristic for all larger architectures and did not tune the hyperparameter any further.
 In global rate distortion regularization (Grdi), the best test error was achieved around $\beta=0.2$ although Grdi seems to behave rather robust in the range $\beta \in [0;0.8]$---see middle left panel in Figure~\ref{fig:results}. 
 In local rate distortion regularization (Lrdi), the best test error was achieved for $\beta=0$ with ordinary utility maximization without regularization---see middle right panel. However, when measuring the performance in terms of expected utility on the test set, Lrdi achieved a significant performance increase compared to ordinary utility maximization in the range $\beta \in [10^{-5},5 \cdot 10^{-4}]$---see upper right panel. In our final studies, we could furthermore ascertain that Lrdi performs reasonably well on larger architectures as it achieved a test error of 1.26\% compared to 1.43\% in ordinary utility maximization when increasing the number of units per hidden layer to 4096.

\begin{table}[h]
\caption{Classification Errors on the MNIST Test Set in the Permutation Invariant Setup}
\label{tab:best_results_non-conv}
\begin{center}
\begin{tabular}{c|c|c}
{Method}  & {$\#_{\text{neu}}$} & {$\text{Error [\%]}$}  \\
\hline 
Bayes by backprop \cite{Blundell2015}           & 1200 & 1.32  \\
Dropout   \cite{Wan2013}                & 800 & 1.28 \\
Dropconnect \cite{Wan2013}                   & 800 & 1.20 \\
Dropout     \cite{Srivastava2014}      & 4096 & \textbf{1.01} \\
\hline 
     & 529 & 1.36  \\
  Local rate distortion (Lrdi)                            & 1024 & 1.34  \\
       $\beta=10^{-5}$        & 2025 & 1.28 \\
                              & 4096 & \textbf{1.26} \\
\hline 
        & 529 & 1.23  \\
   Global rate distortion (Grdi)                           & 1024 & 1.17  \\
        $\beta=0.2$          & 2025 & 1.14 \\
                              & 4096 & \textbf{1.11} \\
   
\end{tabular}
\end{center}
\end{table}
The results of our final studies where we trained networks for 70 epochs are illustrated in Table~\ref{tab:best_results_non-conv} which compares rate distortion regularization to other techniques from the literature for different network architectures comprising two hidden layers. It can be seen that both local and global rate distortion regularization (Lrdi and Grdi respectively) attain results in the permutation invariant setting (Lrdi: 1,26\%, Grdi: 1.11\%) that are competitive with other recent techniques like dropout (1.01\% \cite{Srivastava2014} and 1.28\% \cite{Wan2013}), dropconnect (1.20\% \cite{Wan2013}) and Bayes by backprop (1.32\% \cite{Blundell2015}). It is furthermore shown that both rate distortion regularizers lead to a decreasing generalization error when increasing the number of neurons in hidden layers which demonstrates successful prevention of overfitting. Successful prevention of overfitting is additionally demonstrated by applying  global rate distortion regularization (Grid, $\beta=0.2$) to a convolutional neural network with an architecture according to \cite{Wan2013}---see Section~B.2 in \cite{Wan2013}---attaining an error of 0.61\% without tuning any hyperparameters (see Table~\ref{tab:best_results_conv}). This result is also competitive with other recent techniques in the permutation non-invariant setting---compare to dropout (0.59\% \cite{Wan2013}) and dropconnect (0.63\% \cite{Wan2013}).
In line with \cite{Srivastava2014}, we preprocessed the input with ZCA whitening and added a max-norm regularizer to limit the size of presynaptic weight vectors to at most $3.5$.

\begin{table}[h]
\caption{Classification Errors on the MNIST Test Set in the Permutation Non-Invariant Setup}
\label{tab:best_results_conv}
\begin{center}
\begin{tabular}{c|c}
{Method}  &  {$\text{Error [\%]}$}  \\
\hline 

	Conv net + Dropconnect \cite{Wan2013}  & 0.63 \\
    Conv net + Grdi ($\beta=0.2$)          & 0.61 \\    
    Conv net + Dropout \cite{Wan2013} & \textbf{0.59} \\    
\end{tabular}
\end{center}
\end{table}

The lower panels of Figure~\ref{fig:results} show the development of the test set error over epochs for both rate distortion regularizers (red) compared to ordinary utility maximization without regularization (Umax, black) for the different network architectures that we used in the permutation invariant setting. It can be seen that the global variant of our regularizer (Grdi with $\beta=0.2$, see lower left panel in Figure~\ref{fig:results}) leads to a significant increase in performance across different architectures as demonstrated by the two separate clusters of trajectories. In addition, Grdi also leads to faster learning as the red trajectories in the lower left panel of Figure~\ref{fig:results} decrease significantly faster then the black trajectories during the first ten epochs of training. For the local variant of our regularizer (Lrdi with $\beta=10^{-5}$, see lower right panel in Figure~\ref{fig:results}), the performance improvements are less prominent when compared to the global variant.

\section{CONCLUSION}
\label{sec:conclusion}

Previously, a synaptic weight update rule for a single reward-maximizing spiking neuron was devised, where the neuron was interpreted as a bounded rational decision-maker under limited computational resources with help of rate distortion theory \cite{Leibfried2015}.  It was shown that such a bounded rational weight update rule leads to an efficient regularization by preventing synaptic weights from growing without bounds. In our current work, we extend these results to deterministic neurons and neural networks. On the MNIST benchmark classification task, we have demonstrated the regularizing effect of our approach as networks were successfully prevented from overfitting. These results are robust as we conducted experiments with different network architectures achieving performance competitive with other recent techniques like dropout \cite{Srivastava2014}, dropconnect \cite{Wan2013} and Bayes by backprop \cite{Blundell2015} for both ordinary and convolutional networks. The strength of rate distortion regularization is that it is a more principled approach than for example dropout and dropconnect as it may be applied to general artificial agents with parameterized policies and not only to neural networks.
 Parameterized policies that optimize the rate distortion objective have been previously applied to unsupervised density estimation tasks with autoencoder networks 
   \cite{SanchezGiraldo2013}. 
Our current work extends this kind of approach to the theory of reinforcement and supervised learning with feedforward neural networks, and also provides evidence that this approach scales well on large data sets.

\clearpage

\appendix
\section{APPENDIX}
\label{sec:appendix}

\subsection{MUTUAL INFORMATION RATE OF A DETERMINISTIC NEURON}
\label{sec:mutual_info_rate_det_neu}

\begin{equation}
\label{eq:derivation_mutual_information_rate}
\begin{split}
& \lim_{\Delta t \rightarrow 0} \frac{1}{\Delta t} I(\boldsymbol\xi,y) \\
& =  \lim_{\Delta t \rightarrow 0} \frac{1}{\Delta t} \left\langle  \sum_y p_{\mathbf{w}}(y|\boldsymbol\xi) \ln \frac{p_{\mathbf{w}}(y|\boldsymbol\xi)}{p_{\mathbf{w}}(y)} \right\rangle_{p(\boldsymbol\xi)} \\
& =  \lim_{\Delta t \rightarrow 0} \frac{1}{\Delta t} \left\langle  \phi(\mathbf{w}^\top \boldsymbol\xi) \Delta t \ln \frac{\phi(\mathbf{w}^\top \boldsymbol\xi)}{\bar{\phi}(\mathbf{w})}  \right\rangle_{p(\boldsymbol\xi)}  \\
& + \lim_{\Delta t \rightarrow 0} \frac{1}{\Delta t} \left\langle (1-\phi(\mathbf{w}^\top \boldsymbol\xi) \Delta t) \underbrace{\ln \frac{1-\phi(\mathbf{w}^\top \boldsymbol\xi) \Delta t}{1-\bar{\phi}(\mathbf{w})\Delta t}}_{\rightarrow 0} \right\rangle_{p(\boldsymbol\xi)} \\
& =  \left\langle \phi(\mathbf{w}^\top \boldsymbol\xi) \ln \frac{\phi(\mathbf{w}^\top \boldsymbol\xi)}{\bar{\phi}(\mathbf{w})} \right\rangle_{p(\boldsymbol\xi)} .
\end{split}
\end{equation}

\subsection{DERIVATIVE OF THE MUTUAL INFORMATION RATE}
\label{sec:derivative_mutual_info_rate}

\begin{equation}
\label{eq:derivative_mutual_information_rate}
\begin{split}
& \frac{\partial}{\partial w_i} \lim_{\Delta t \rightarrow 0} \frac{1}{\Delta t} I(\boldsymbol\xi,y) \\
& = \lim_{\Delta t \rightarrow 0} \frac{1}{\Delta t} \frac{\partial}{\partial w_i} \left\langle \sum_y p_{\mathbf{w}}(y|\boldsymbol\xi) \ln \frac{p_{\mathbf{w}}(y|\boldsymbol\xi)}{p_{\mathbf{w}}(y)} \right\rangle_{p(\boldsymbol\xi)} \\
& = \lim_{\Delta t \rightarrow 0} \frac{1}{\Delta t}  \left\langle \sum_y \left( \frac{\partial}{\partial w_i}  p_{\mathbf{w}}(y|\boldsymbol\xi) \right) \ln \frac{p_{\mathbf{w}}(y|\boldsymbol\xi)}{p_{\mathbf{w}}(y)} \right\rangle_{p(\boldsymbol\xi)} \\
& + \lim_{\Delta t \rightarrow 0} \frac{1}{\Delta t}  \underbrace{ \left\langle \sum_y p_{\mathbf{w}}(y|\boldsymbol\xi) \left( \frac{\partial}{\partial w_i} \ln p_{\mathbf{w}}(y|\boldsymbol\xi) \right) \right\rangle_{p(\boldsymbol\xi)} }_{= \left\langle \sum_y \frac{\partial}{\partial w_i} p_{\mathbf{w}}(y|\boldsymbol\xi) \right\rangle_{p(\boldsymbol\xi)} = \frac{\partial}{\partial w_i} 1 = 0} \\
& - \lim_{\Delta t \rightarrow 0} \frac{1}{\Delta t}  \underbrace{\left\langle \sum_y p_{\mathbf{w}}(y|\boldsymbol\xi) \left( \frac{\partial}{\partial w_i} \ln p_{\mathbf{w}}(y) \right) \right\rangle_{p(\boldsymbol\xi)} }_{ = \sum_y \frac{\partial}{ \partial w_i} p_{\mathbf{w}}(y) = \frac{\partial}{ \partial w_i} 1 = 0} \\
& = \lim_{\Delta t \rightarrow 0} \frac{1}{\Delta t}  \left\langle  \xi_i \phi'(\mathbf{w}^\top \boldsymbol\xi) \Delta t \ln \frac{\phi(\mathbf{w}^\top \boldsymbol\xi)}{\bar{\phi}(\mathbf{w})} \right\rangle_{p(\boldsymbol\xi)} \\
& - \lim_{\Delta t \rightarrow 0} \frac{1}{\Delta t}  \left\langle  \xi_i \phi'(\mathbf{w}^\top \boldsymbol\xi) \Delta t \underbrace{\ln \frac{1-\phi(\mathbf{w}^\top \boldsymbol\xi) \Delta t}{1-\bar{\phi}(\mathbf{w})\Delta t}}_{\rightarrow 0} \right\rangle_{p(\boldsymbol\xi)} \\
& = \left\langle \xi_i \phi'(\mathbf{w}^\top \boldsymbol\xi) \ln \frac{\phi(\mathbf{w}^\top \boldsymbol\xi)}{\bar{\phi}(\mathbf{w})} \right\rangle_{p(\boldsymbol\xi)} .
\end{split}
\end{equation}

\subsection{DERIVATIVE OF THE GLOBAL MUTUAL INFORMATION}
\label{sec:derivative_mutual_info_global}

\begin{equation}
\label{eq:derivative_mutual_information_global}
\begin{split}
& \frac{\partial}{\partial w^n_i} I(\boldsymbol\xi,\mathbf{y}) \\
& = \frac{\partial}{\partial w^n_i} \left\langle \sum_\mathbf{y} p_{\mathcal{W}}(\mathbf{y}|\boldsymbol\xi) \ln \frac{p_{\mathcal{W}}(\mathbf{y}|\boldsymbol\xi)}{p_{\mathcal{W}}(\mathbf{y})} \right\rangle_{p(\boldsymbol\xi)} \\
& = \left\langle \sum_\mathbf{y} \left( \frac{\partial}{\partial w^n_i} p_{\mathcal{W}}(\mathbf{y}|\boldsymbol\xi) \right) \ln \frac{p_{\mathcal{W}}(\mathbf{y}|\boldsymbol\xi)}{p_{\mathcal{W}}(\mathbf{y})} \right\rangle_{p(\boldsymbol\xi)} \\
& +  \underbrace{ \left\langle \sum_\mathbf{y} p_{\mathcal{W}}(\mathbf{y}|\boldsymbol\xi) \left( \frac{\partial}{\partial w^n_i} \ln p_{\mathcal{W}}(\mathbf{y}|\boldsymbol\xi) \right) \right\rangle_{p(\boldsymbol\xi)}}_{= \left\langle \sum_{\mathbf{y}} \frac{\partial}{\partial w^n_i} p_{\mathcal{W}}(\mathbf{y}|\boldsymbol\xi) \right\rangle_{p(\boldsymbol\xi)} = \frac{\partial}{\partial w^n_i} 1 = 0 }  \\
& - \underbrace{ \left\langle \sum_\mathbf{y} p_{\mathcal{W}}(\mathbf{y}|\boldsymbol\xi) \left( \frac{\partial}{\partial w^n_i} \ln p_{\mathcal{W}}(\mathbf{y}) \right) \right\rangle_{p(\boldsymbol\xi)}}_{ = \sum_{\mathbf{y}} \frac{\partial}{\partial w^n_i} p_{\mathcal{W}}(\mathbf{y}) = \frac{\partial}{\partial w^n_i} 1 = 0} \\
& = \left\langle \sum_j \left( \frac{\partial}{\partial w^n_i} f_j(\mathcal{W},\boldsymbol\xi) \right) \ln \frac{f_j(\mathcal{W},\boldsymbol\xi)}{\bar{f}_j(\mathcal{W})} \right\rangle_{p(\boldsymbol\xi)} .
\end{split}
\end{equation}

\subsubsection*{Acknowledgements}

This study was supported by the DFG, Emmy Noether grant BR4164/1-1.


\subsubsection*{References}

\renewcommand{\section}[2]{}
\bibliographystyle{unsrt}
\bibliography{refs}

\end{document}